\documentclass[sigconf]{acmart}

\usepackage{booktabs} 
\usepackage{balance} 

\usepackage{subcaption}

\usepackage[english]{babel}
\usepackage[utf8x]{inputenc}
\usepackage[T1]{fontenc}

\usepackage{amsmath}
\usepackage{graphicx}
\usepackage[colorinlistoftodos]{todonotes}

\newcommand{\el}{\textit{et al.}}

\title[Evolving Mario Levels in the Latent Space of a DCGAN]{Evolving Mario Levels in the Latent Space of a Deep Convolutional Generative Adversarial Network}

\author[V. Volz]{Vanessa Volz}
\affiliation{%
 \institution{TU Dortmund University}
 \city{Dortmund, Germany} 
}
\email{vanessa.volz@tu-dortmund.de}

\author[J. Schrum]{Jacob Schrum}
\orcid{0000-0002-7315-0515} 
\affiliation{%
 \institution{Southwestern University}
 \city{Georgetown} 
 \state{TX USA} 
 \postcode{78626 USA}
}
\email{schrum2@southwestern.edu}

\author[J. Liu]{Jialin Liu}
\affiliation{%
 \institution{Queen Mary University of London}
 \city{London, UK} 
}
\email{jialin.liu@qmul.ac.uk}

\author[S. M. Lucas]{Simon M. Lucas}
\affiliation{%
 \institution{Queen Mary University of London}
 \city{London, UK} 
}
\email{simon.lucas@qmul.ac.uk}

\author[A. Smith]{Adam Smith}
\affiliation{%
 \institution{University of California}
 \city{Santa Cruz, CA USA} 
}
\email{amsmith@ucsc.edu}

\author[S. Risi]{Sebastian Risi}
\affiliation{%
 \institution{IT University of Copenhagen}
 \city{Copenhagen, Denmark} 
}
\email{sebr@itu.dk}

\copyrightyear{2018}
\acmYear{2018}
\setcopyright{acmlicensed}
\acmConference[GECCO '18]{Genetic and Evolutionary Computation Conference}{July 15--19, 2018}{Kyoto, Japan}
\acmBooktitle{GECCO '18: Genetic and Evolutionary Computation Conference, July 15--19, 2018, Kyoto, Japan}
\acmPrice{15.00}
\acmDOI{10.1145/3205455.3205517}
\acmISBN{978-1-4503-5618-3/18/07}

\begin{document}

\begin{abstract}
Generative Adversarial Networks (GANs) are a machine learning approach capable of generating novel example outputs across a space of provided training examples. Procedural Content Generation (PCG) of levels for video games could benefit from such models, especially for games where there is a pre-existing corpus of levels to emulate. This paper trains a GAN to generate levels for Super Mario Bros using a level from the Video Game Level Corpus. The approach successfully generates a variety of levels similar to one in the original corpus, but is further improved by application of the Covariance Matrix Adaptation Evolution Strategy (CMA-ES). Specifically, various fitness functions are used to discover levels within the latent space of the GAN that maximize desired properties. Simple static properties are optimized, such as a given distribution of tile types. Additionally, the champion A* agent from the 2009 Mario AI competition is used to assess whether a level is playable, and how many jumping actions are required to beat it. These fitness functions allow for the discovery of levels that exist within the space of examples designed by experts, and also guide the search towards levels that fulfill one or more specified objectives.

\end{abstract}

\keywords{Generative Adversarial Network, Procedural Content Generation, Mario, CMA-ES, Game}

\maketitle

\section{Introduction}\label{sec:intro}

Procedural Content Generation (PCG) covers the creation of game content (e.g., game rules, levels, characters, background stories, textures and sound) by algorithms with or without help from human designers~\cite{togelius2011procedural}. 
The history of digital PCG goes back to the 1980s, when the game \emph{Elite}\footnote{\url{https://en.wikipedia.org/wiki/Elite_(video_game)}} was published. Due to the limited memory capacities of personal computers of the time, a decision was made to save only the seed to a random generation process rather than to store complete level designs. From a specified seed value, a generator would proceed to deterministically (pseudo-randomly) recreate a sequence of numbers which were then used to determine the names, positions, and other attributes of game objects.
The adoption of PCG exploded during the 2000s when it was picked up in application to game graphics \cite{Ebert:2002:TMP:572337}.
Since then, much work has sprung up around PCG in both the industry and academic spheres~\cite{shaker2016procedural}. Additionally, various competitions have been organized in international conferences during recent years, such as the Mario AI Level Generation 
Competition\footnote{\url{http://www.marioai.org/LevelGeneration}}, 
Platformer AI Competition\footnote{\url{https://sites.google.com/site/platformersai/LevelGeneration}}, AI Birds Level Generation Competition\footnote{\url{https://aibirds.org/other-events/level-generation-competition.html}} and the General Video Game AI (GVGAI)\footnote{\url{http://www.gvgai.net/}} Level Generation Competition~\cite{khalifa2016general}. The approach introduced here is an example of PCG via Machine Learning (PCGML; \cite{summerville2017procedural}), which is a recently emerging research area.

\begin{figure}
\centering
\includegraphics[width=.5\textwidth]{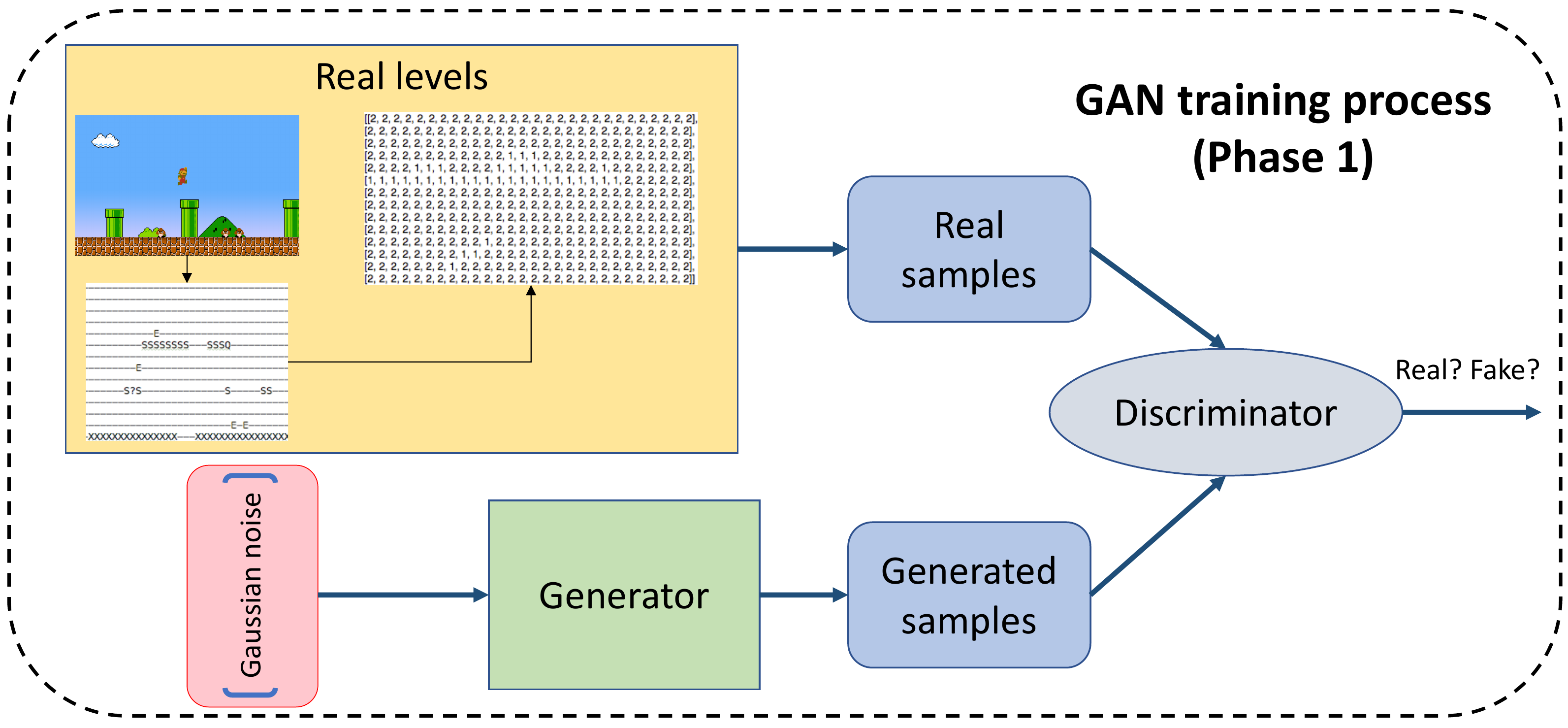}
\includegraphics[width=.5\textwidth]{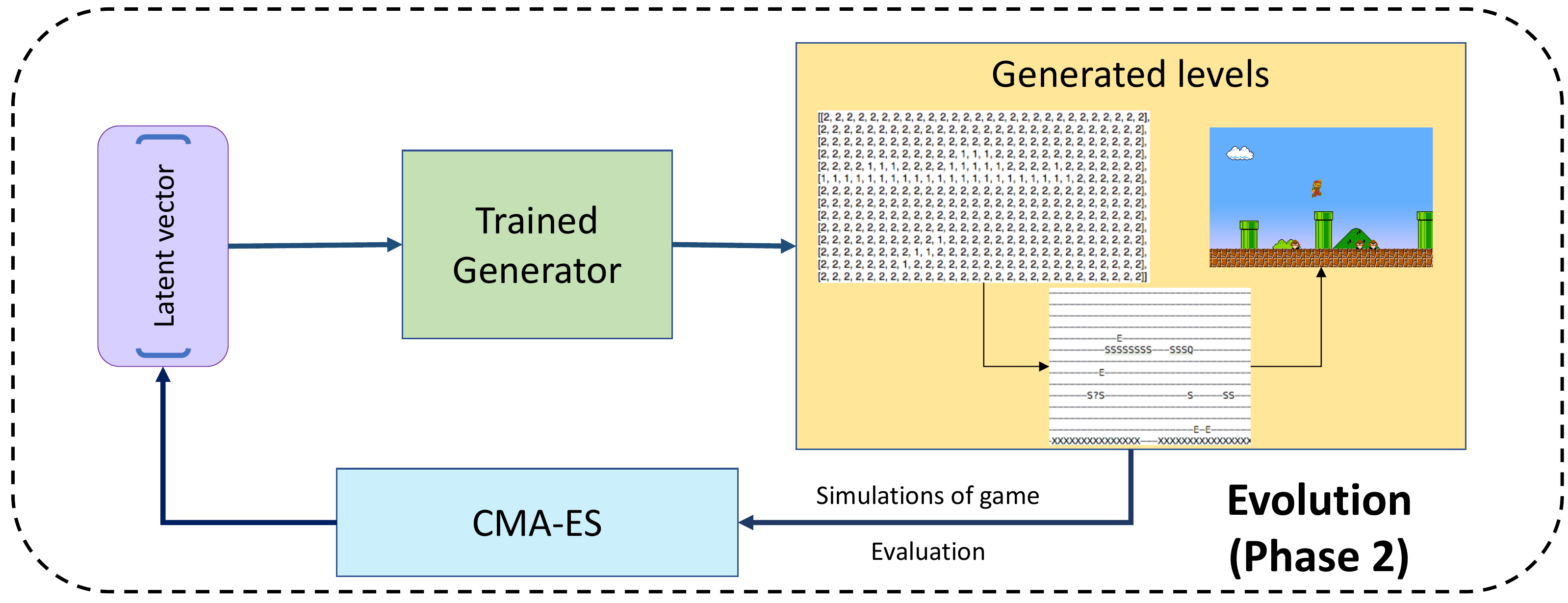}
\caption{Overview of the GAN training process and the evolution of latent vectors. \normalfont The approach is divided into two distinct phases. In Phase 1 a GAN is trained in an unsupervised way to generate Mario levels. In the second phase, we search for latent vectors that produce levels with specific properties. }
\label{fig:training}
\end{figure}
The approach presented in this paper is to create new game levels that emulate those designed by experts using a variant of a Generative Adversarial Network (GAN)~\cite{goodfellow2014generative}. GANs are deep neural networks trained in an unsupervised 
way that have shown exceptional promise in reproducing aspects of images from a training set.  Additionally, the space of levels encoded by the GAN is further searched using the Covariance Matrix Adaptation Evolutionary Strategy (CMA-ES)~\cite{hansen2003reducing}, in order to discover levels with particular attributes. The idea of \emph{latent variable evolution} (LVE)  was recently introduced in the context of interactive evolution of images~\cite{bontrager2018deep} and fingerprint generation~\cite{bontrager2017deepmasterprint}  but so far has not been applied to PCG of video game levels. 

The specific game in this paper is \emph{Super Mario Bros}\footnote{\url{https://en.wikipedia.org/wiki/Super_Mario_Bros.}},
but the technique should generalize to any game for which an existing corpus of levels is available.
Our GAN is trained on a single level from the original
\emph{Super Mario Bros}, available as part of the 
Video Game Level Corpus (VGLC)~\cite{Summerville:pcg2016-VGLC}. 
CMA-ES is then used to find ideal inputs to the GAN from within its latent vector space (Figure~\ref{fig:training}). During the evolution, the generated levels are evaluated using different fitness functions. This allows for the discovery of levels that exist between and beyond those sparse examples designed by human designers, and that also optimize additional goals. Our approach is capable of generating playable levels that meet various goals and is ready to be applied to level generation of other games, such as the games in the GVGAI framework. By training on only a single level, we are able to show that even with a very limited dataset, we can apply the presented approach successfully.

The rest of this paper is structured as follows. Section~\ref{sec:back} introduces the background and related work. The main approach is described in Section~\ref{sec:approach}. Section~\ref{sec:xp} details the experimental design. The experimental results are presented and discussed in Section~\ref{sec:res}. Section~\ref{sec:conc} then concludes the paper.


\section{Background and Related Work}\label{sec:back}
In this section, Procedural Content Generation for games is discussed, followed descriptions of technical tools applied in this paper: GANs, latent variable evolution, and CMA-ES. 

\subsection{Procedural content generation}
Togelius~\el~\cite{togelius2011procedural} defined Procedural Content Generation (PCG) as the \emph{algorithmic creation of game content with limited or indirect user input}~\cite{togelius2011procedural,togelius2011search,togelius2013procedural}.
Examples of game content include game rules, levels, maps/mazes, characters, weapons, vehicles, background stories, textures and sound. Automatic game level generation, with little or no human intervention, is a challenging problem. For some games, the levels are represented as maps or mazes~\cite{gazzard2013mazes}. Examples include \emph{Doom}, \emph{Pac-Man}, and \emph{Super Mario Bros}, one of the classic platform video games created by \emph{Nintendo}.

The first academic Procedural Content Generation competition was the 2010 Mario AI Championship~\cite{shaker20112010}, in which the participants were required to submit a level generator which implements a provided Java interface and returns a new level within $60$ seconds. The competition framework was implemented based on \emph{Infinite Mario Bros}~\footnote{\url{https://tinyurl.com/yan4ep7g}}, a public clone of \emph{Super Mario Bros}.

The availability and popularity of the Mario AI framework
has led to several approaches for generating levels for \emph{Super Mario Bros}.
Shaker~\el~\cite{shaker2012evolving} evolved Mario levels using Grammatical Evolution (GE).
In 2016, Summerville and Mateas~\cite{summerville2016super} applied Long Short-Term Memory Recurrent Neural Networks (LSTMs) to generate game levels trained on existing Mario levels, and then improved the generated levels by incorporating player path information. This approach inspired a novel approach to level generation, in which new levels are generated automatically from a sketch of some desired path drawn by a human designer. Another approach that was trained using existing Mario levels is that of Jain~\el~\cite{jain2016autoencoders}, which trained auto-encoders to generate new levels using a binary encoding where empty (accessible) spaces are represented by 0 and the others (e.g., terrain, enemy, tunnel, etc.) by 1. Though this approach could generate
interesting levels, the use of random noise inputs into the trained auto-encoder sometimes led to problematic levels.
Additionally, because of the binary encoding, no distinction was made between various possible types of tiles.


\subsection{Generative Adversarial Networks}
\label{sec:GAN}

Generative Adversarial Networks (GANs) were first introduced by Goodfellow~\el~\cite{goodfellow2014generative} in 2014. Their training process can be seen as a two-player adversarial game in which a generator $G$ (faking samples decoded from a random noise vector) and a discriminator $D$ (distinguishing real/fake samples and outputting 0 or 1) are trained at the same time by playing against each other. The discriminator $D$ aims at minimizing the probability of mis-judgment, while the generator $G$ aims at maximizing that probability. Thus, the generator is trained to deceive the discriminator by generating samples that are good enough to be classified as genuine. Training ideally reaches a steady state where $G$ reliably generates realistic examples and $D$ is no more accurate than a coin flip.

GANs quickly became popular in some sub-fields of computer vision, such as image generation. However, training GANs is not trivial and often results in unstable models. Many extensions have been proposed, such as Deep Convolutional Generative Adversarial Networks (DCGANs)~\cite{radford2015unsupervised}, a class of Convolutional Neural Networks (CNNs); Auto-Encoder Generative Adversarial Networks (AE-GANs)~\cite{makhzani2015adversarial}; and Plug and Play Generative Networks (PPGNs)~\cite{nguyen2016plug}.
A particularly interesting variation are Wasserstein GANs (WGANs)~\cite{arjovsky2017wasserstein,gulrajani2017improved}. WGANs minimize the approximated Earth-Mover (EM) distance (also called Wasserstein metric), which is used to measure how different the trained model distribution and the real distribution are. WGANs have been demonstrated to achieve more stable training than standard GANs.

At the end of training, the discriminator $D$ is discarded, and the generator $G$ is used to produce new, novel outputs that capture the fundamental properties present in the training examples. The input to $G$ is some fixed-length vector from a latent space (usually sampled from a block-uniform or isotropic Gaussian distribution). For a properly trained GAN, randomly sampling vectors from this space should produce outputs that would be mis-classified as examples of the target class with equal likelihood to the true examples. However, even if all GAN outputs are perceived as valid members of the target class, there could still be a wide range of meaningful variation within the class that a human designer would want to select between. A means of searching within the real-valued latent vector space of the GAN would allow a human to find members of the target class that satisfy certain requirements.

\subsection{Latent variable evolution}

The first \emph{latent variable evolution} (LVE) approach was introduced by Bontrager~\el~\cite{bontrager2017deepmasterprint}. In their work the authors train a GAN on a set of real fingerprint images and then apply evolutionary search to find a latent vector that matches with as many subjects in the dataset as possible. 
\begin{figure*}[t]
\centering
\includegraphics[width=1.0\textwidth]{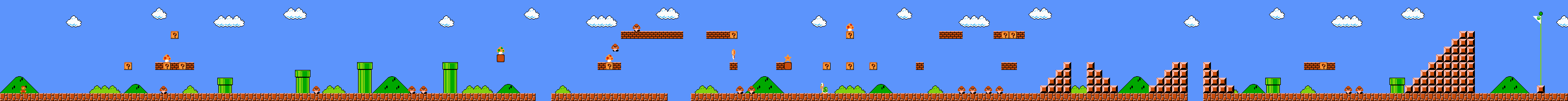}
\caption{The Training Level. \normalfont The training data is generated  
by sliding a 28 $\times$ 14  window over the level from left to right, one tile at a time.}
\label{fig:trainingsample}
\end{figure*}

In another paper Bontrager et al.~\cite{bontrager2018deep} present an interactive evolutionary system, in which users can evolve the latent vectors for a GAN trained on different classes of objects (e.g.\ faces or shoes). Because the  GAN is trained on a specific target domain, it becomes a compact and robust genotype-to-phenotype mapping (i.e. most produced phenotypes do resemble valid domain artifacts) and users were able to guide evolution towards images that closely resembled given target images. Such target based evolution has been shown to be challenging with other indirect encodings \cite{woolley2011deleterious}.

Because of the promising previous LVE approaches, in this paper we investigate how latent GAN vectors can be evolved through a fitness-based approach in the context of level generation. 

\subsection{CMA-ES}

Covariance Matrix Adaptation Evolutionary Strategy (CMA-ES)~\cite{hansen2003reducing} is a powerful and widely used evolutionary algorithm that is well suited for evolving vectors of real numbers. The CMA-ES is a second-order method using the covariance matrix estimated iteratively by finite differences. It has been demonstrated to be efficient for optimizing non-linear non-convex problems in the continuous domain without a-priori domain knowledge, and it does not rely on the assumption of a smooth fitness landscape.

We applied CMA-ES to evolve the latent vector and applied several fitness functions on the generated levels. Fitness functions can be based on purely static properties of the generated levels, or on the results of game simulations using artificial agents.

\section{Approach}\label{sec:approach}

The approach is divided into two main phases, visualised in Figure~\ref{fig:training}. First, a GAN is trained on an existing Mario level (Figure~\ref{fig:trainingsample}). The level is encoded as a multi-dimensional array as described in Section \ref{sec:levelrep} and depicted in the yellow box. The generator (green) operates on a Gaussian noise vector (red) and is trained to output levels using the same representation. The discriminator is then employed to tell the existing and generated levels apart. Both the generator and discriminator are trained using an adversarial learning process as described in Section \ref{sec:GAN}.

Once this process is completed, the generator network of the GAN, $G$, can be viewed as our learned genotype-to-phenotype mapping that takes as input a latent vector (blue) of real numbers (of size 32 in the experiments in this paper) and produces a tile-level description of a Mario level. Instead of simply drawing independent random samples from the latent space, we put exploration under evolutionary control (using a CMA-ES in this case). In other words, we search through the space of latent vectors to produce levels with different desirable properties such as distributions of tiles, difficulty, etc. Specific parts of the training process are discussed in the following.

\subsection{Level representation}\label{sec:levelrep}

Mario levels have different representations within the Video Game Level Corpus (VGLC)~\cite{Summerville:pcg2016-VGLC} and Mario AI framework.
Both representations are tile based.
Specifically, each Mario level from the VGLC
uses a particular character symbol to represent each possible
tile type. However, it should be noted that this VGLC representation
is primarily concerned with functional properties of tiles
rather than artistic properties, and is thus incapable of
distinguishing certain visually distinct tile types. The 
only exception are \emph{pipes}, which are represented by
four visually distinct tile types, despite all being functionally
equivalent to an impassable ground block. Interestingly,
the VGLC encoding ignores functional differences between
different enemy types by providing only a single character
symbol to represent enemies, which we choose to map to the
generic \emph{Goomba} enemy type. 


To encode the levels for training,
each tile type is represented by
a distinct integer, which is converted to a one-hot
encoded vector before being input into the
discriminator. 
The generator network also outputs
levels represented using 
the one-hot encoded format, which is then converted
back to a collection of integer values.
Levels in this integer-based format are then sent to
the Mario AI framework for rendering.
Mario AI allows for a
broader range of artistic diversity in its tile types,
but because of the simplicity of the VGLC encoding,
only a simple subset of the available Mario AI tiles
are used. The mapping from VGLC tile types and symbols, 
to GAN training number codes, and finally to Mario AI 
tile visualizations
is detailed in the Table~\ref{tab:tiles}. 

\begin{table}
\centering
\caption{\label{tab:tiles}Tile types used in generated Mario levels. \normalfont 
The symbol characters come from the VGLC encoding, and the
numeric identity values are then mapped to the corresponding values
employed by the Mario AI framework to produce the visualization shown.
The numeric identity values are 
expanded into one-hot vectors when input into the
discriminator network during GAN training.}
\begin{tabular}{cccc}
\hline
Tile type & Symbol & Identity & Visualization\\
\hline 
Solid/Ground & X & 0 & \includegraphics[scale=0.5]{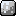}\\
Breakable & S & 1 & \includegraphics[scale=0.5]{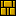}\\
Empty (passable) & - & 2 & \\
Full question block & ? & 3 & \includegraphics[scale=0.5]{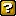}\\
Empty question block & Q & 4 & \includegraphics[scale=0.5]{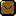}\\
Enemy & E & 5 & \includegraphics[scale=0.5]{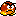}\\
Top-left pipe & < & 6 & \includegraphics[scale=0.5]{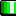}\\
Top-right pipe & > & 7 & \includegraphics[scale=0.5]{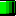}\\
Left pipe & [ & 8 & \includegraphics[scale=0.5]{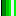}\\
Right pipe & ] & 9 & \includegraphics[scale=0.5]{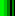}\\
\hline
\end{tabular}
\end{table}


The GAN input files were created by processing a level file 
from the VGLC for the original Nintendo game \emph{Super Mario Bros}, which is shown in Figure~\ref{fig:trainingsample}.
Each level file is a plain text file where each line of the file corresponds to a row
of tiles in the Mario level.  Within a level all rows are of the same length,
and each level is 14 tiles high.  The GAN expected to always see a rectangular
image of the same size, hence each input image was generated
by sliding a 28 (wide) x 14 (high)
window over the raw level from left to right, one tile at a time. The width of 28 tiles is equal to the width of the screen in Mario. 
 In the input files each tile type is
represented by a specific character, which was then mapped to a specific integer in the training
images, as listed in Table~\ref{tab:tiles}.  
This procedure created a set of 173 training images. 

While we could have used a larger dataset instead of this relatively small one, its use allows us to test the GAN's ability to learn from relatively little data, 
which could be especially important for games that do not offer such a large training corpus as Mario. Additionally, because of the smaller  training set it is possible to manually inspect if the LVE approach is able to generate levels with properties not directly  found in the training set itself. 

\subsection{GAN training}
\begin{figure}[h]
\centering
\includegraphics[width=\columnwidth]{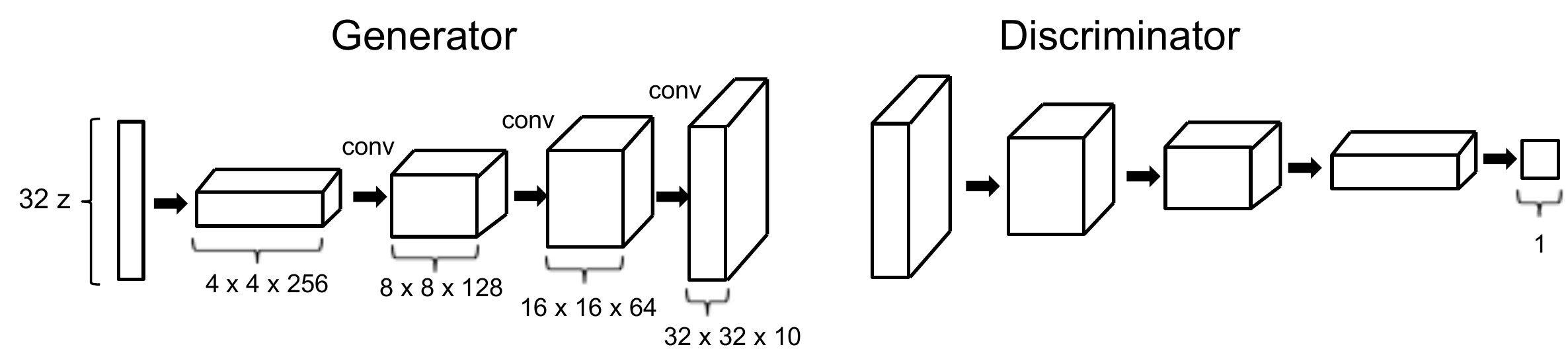}
\caption{The Mario DCGAN architecture.}
\label{fig:architecture}
\end{figure}

Our Deep Convolutional GAN (DCGAN) is adapted from the model in \cite{arjovsky2017wasserstein} and trained with the WGAN algorithm. 
The network architecture is shown in Figure~\ref{fig:architecture}. Following the original DCGAN architecture, the network uses strided convolutions in the discriminator and fractional-strided convolutions in the generator.  Additionally, we employ \emph{batchnorm} both in the generator and discriminator after each layer. 
In contrast to the original architecture in \cite{arjovsky2017wasserstein}, we use ReLU activation functions for all layers in the generator, even for the output (instead of Tanh), which we found gave better results. Following \cite{arjovsky2017wasserstein}, the discriminator uses LeakyReLU activation in all layers. 

When training the GAN, each integer tile was expanded to a one-hot vector.
Therefore the training inputs for the discriminator are 10 channels (one-hot across 10 possible tile types) of size 32 $\times$ 32 (the DCGAN implementation we used required the input size to be a multiple of 16 so the levels were padded). For example, in the first channel, the location of ground titles are marked with a 1.0, while all other locations are set to 0.0. The size of the latent vector input to the generator has a length of 32. 

Once training of the GAN is completed the generator represents our learned genotype-to-phenotype mapping. When running evolution, the final $10 \times 32 \times 32$ dimensional output of this generator is cropped to $10 \times 28 \times 14 $ and each output vector for a tile is converted to an integer using the argmax operator, resulting in a level that can be decoded by the Mario AI framework. 



\section{Experiments}\label{sec:xp}
The approach of this paper is tested in two different sets of experiments that can be divided into representation-based and agent-based testing, which are described in more detail below. The experiments are intended as a proof of concept. To apply the proposed approach within a game, the employed fitness functions need to be designed more carefully to correspond to the intended purpose and required properties of the generated content.
The whole project is available on Github\footnote{\url{https://github.com/TheHedgeify/DagstuhlGAN}}.

\subsection{Representation-based testing}
In the representation-based scenarios we directly optimize for a certain distribution of tiles using CMA-ES. In more detail, we test (1) if the approach can generate levels with a certain number of ground titles, and (2) a combination of ground titles and number of enemies. The goal of the second experiment is to create a level composed of multiple subsections that increases gradually in difficulty. 
In all experiments, we seek to minimize the following functions.


Fitness in the first experiment is based on the distance between the produced fraction of ground titles $g$ and the targeted fraction $t$:
\[ 
F_{ground} = \sqrt{ (g-t)^2 }.
\]
In the second experiment, we evolve five different subsections with 100\% ground coverage for sections 1 and 2, and 70\% coverage for sections 3--5. For the fourth and fifth subsection fitness is also determined by  maximizing the total number of $n$ of enemies:
\[
F = F_{ground} + 0.5 (20.0 - n).
\]
This particular weighting was found through prior experimentation. 



\subsection{Agent-based testing}

While being able to generate levels with exactly the desired number of ground tiles and enemies is one desirable feature of a level generator, a fitness function based entirely on the level representation has two inherent weaknesses:
\begin{itemize}
\item Levels with maximal fitness value might not be playable, especially if they are optimized for a small number of ground tiles and/or a large number of enemies.
\item The number of ground tiles and enemies does not necessarily affect the playthrough of a human or AI agent, and may thus not result in levels with the desired difficulty. E.g., the enemies might fall into a hole before Mario can reach them or there might exist an alternative route that avoids difficult jumps.
\end{itemize}
These problems can be alleviated by using an evaluation that is based on playthrough data instead of just the level representation. This way, playability can be explicitly tested and characteristics of a playthrough can be observed directly.

To this end, we implemented agent-based testing using the Mario AI competition framework, as there are a variety of agents already available \cite{togelius2013championship}. 
The CMA-ES is used to find levels that optimize an agent-based fitness function described in the following.
To evaluate a level, the latent vector in question is mapped to $[-1,1]^n$ with a sigmoid function and then sent to the generator model in order to obtain the corresponding level. The level is then imported into the Mario AI framework using the encoding detailed in Table \ref{tab:tiles}, so that agent simulations can be run.

While there are a variety of properties that can be measured using agent-based testing, for this proof-of-concept we chose to specifically focus on the two weaknesses of representation-based fitness functions mentioned above. As before, our use case is to find playable levels with a scalable difficulty.
\begin{figure*}[t]
\centering
	\begin{subfigure}[b]{1.0\textwidth} 
\includegraphics[width=1.0\textwidth]{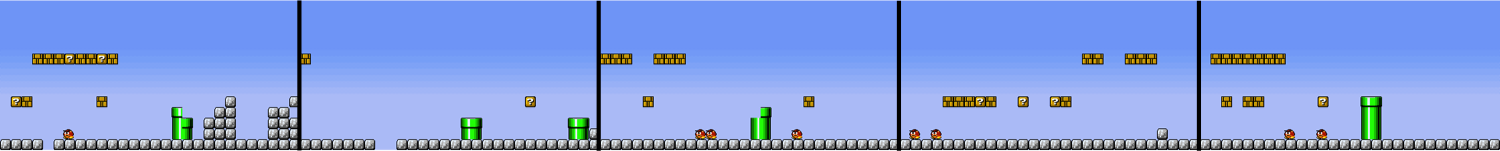}
\caption{Random Sampling}
    \label{fig:examples:rand}
\vspace{0.1in}
\end{subfigure}
	\begin{subfigure}[b]{1.0\textwidth}
\includegraphics[width=1.0\textwidth]{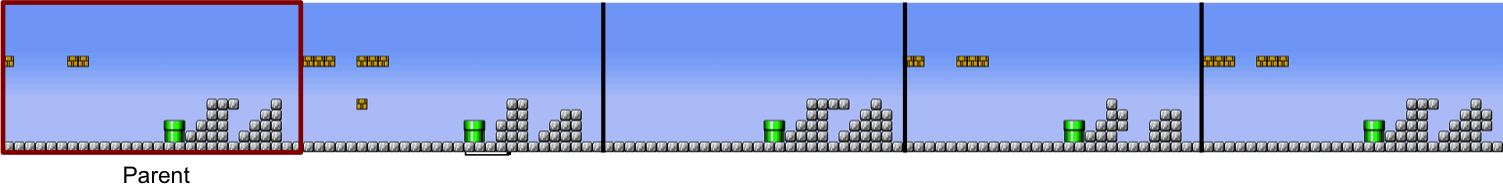}\vspace{-0.1in}
\caption{Mutations}
\label{fig:examples:mut}
\end{subfigure}
\caption{Generated Examples. \normalfont Shown are samples produced by the GAN by (a) sampling random latent vectors, and (b) randomly mutating a specific latent vector. The main result is that the generator is able to produce a wide variety of different level layouts, but varied offspring still resemble their parent.}
\label{fig:gaussian}
\end{figure*}

Given that the A* agent by Robin Baumgarten\footnote{\url{https://www.youtube.com/watch?v=DlkMs4ZHHr8}} (winner of the 2009 Mario AI competition) performs at a super-human level, we use its performance to determine the playability of a given level. For an approximation of experienced difficulty, we use the number of jump actions performed by the agent. The correlation between the number of jumps and difficulty is an assumption, however, jumping is the main mechanic in Mario and is required to overcome obstacles such as holes and enemies. The fitness function we seek to minimize is:
\begin{align*} 
	F_{1} = \begin{cases} - p & \text{for } p<1\\
	- p - \#jumps & \text{for } p=1,
	\end{cases}
\end{align*} where $p$ is the fraction of the level that was completed in terms of progress on the x-axis.

In order to investigate the controllability of the level generation process via agent-based testing, we ran additional experiments where we sought playable levels with a minimal number of required jumps. The fitness function in this case is
\begin{align*} 
	F_{2} = \begin{cases} - p + 60 & \text{for } p<1\\
	- p + \#jumps & \text{for } p=1,
	\end{cases}
\end{align*} where $p$ is the fraction of the level that was completed in terms of progress on the x-axis. The offset of $60$ for the incomplete levels was chosen after preliminary experiments so that unbeatable levels where the agent is trapped and repeatedly jumps are discouraged. As a result,
passable levels will always score a higher fitness than impassable ones.


Since the exact number of jumps is non-deterministic and can produce outliers if the agent gets stuck under an overhang, the actual fitness value in both cases is the average of 10 simulations.









\subsection{Experimental parameters}
For the non-agent testing we use a Python CMA-ES implementation\footnote{\url{https://pypi.python.org/pypi/cma}}.
Because Mario AI is implemented in Java, we use a Java implementation of CMA-ES for the agent-based testing\footnote{\url{https://www.lri.fr/~hansen/cmaes_inmatlab.html\#java}} to evolve the latent vector passed to the trained Python generator. 

For both Java and Python, the CMA-ES population size is $\lambda=14$. For the non-agent based setting we set the initial point to 0, while we set it to a random point within $[-1,1]^n$ for the more complex fitness function in the agent-based setting. Similarly, the standard deviation is initialized to $1.0$ for non-agent and $2.0$ for agent-based testing. The CMA-ES is run for a maximum of $1,000$ function evaluations. 

A total of 20 runs were performed for the non-agent based experiments and 40 runs each for both ($F_{1}$ and $F_{2}$) of the agent-based  CMA-ES experiments.  

Our WGAN implementation is built on a modified version of the original PyTorch WGAN code\footnote{\url{https://github.com/martinarjovsky/WassersteinGAN}}. Both the generator and discriminator are trained with RMSprop with a batch size of 32 and the default  learning rate of 0.00005 for 5,000 iterations. 

\section{Results}\label{sec:res}
To get a better understanding of the GAN's suitability as a genotype-to-phenotype mapping we first tested for expressivity of the encoding and to what degree it has locality (i.e.\ small mutations resulting in small phenotype changes). Figure~\ref{fig:gaussian} shows examples of (a) a randomly sampled GAN and (b) samples around a particular latent vector generated by adding uniformly sampled noise in the range [-0.3, 0.3]. While some aspects (e.g.\ pipes) are sometimes not captured perfectly, the GAN is able to generate a variety of different level layouts that capture some important aspects of the training corpus (Figure~\ref{fig:gaussian}). Additionally, mutations around a particular latent vector vary in interesting ways while still resembling the parent vector. 

\subsection{Representation-based testing}
\begin{figure}[htbp]
\centering
\includegraphics[width=.5\textwidth]{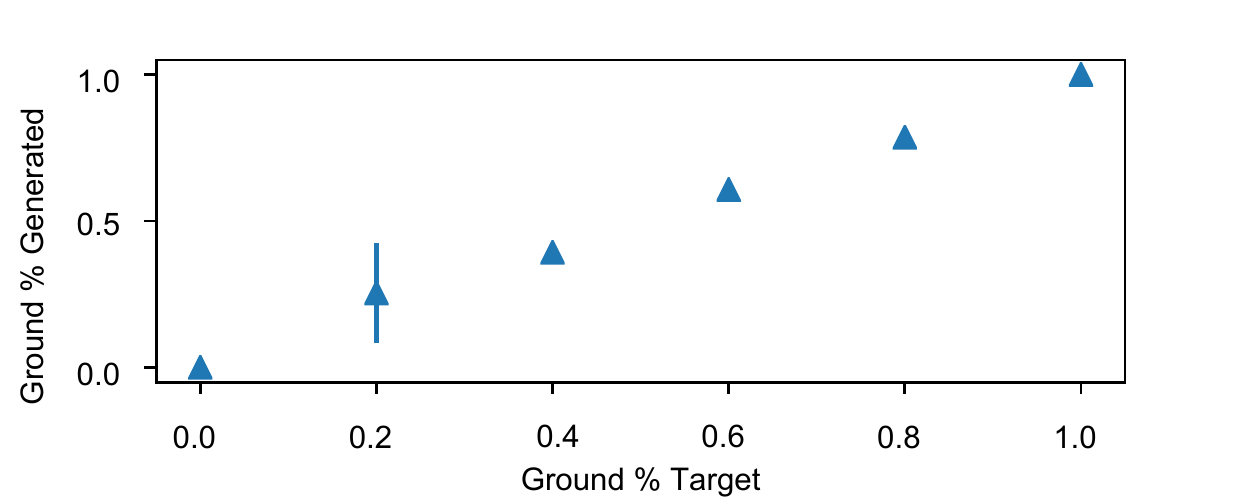}
\caption{Optimized for different percentage of ground tiles. \normalfont Mean values across 20 runs are shown along with one standard deviation. Except for a ground level fraction of 20\% the approach is able to always discover the latent code that produces the desired target percentage of ground tiles. }
\label{fig:ground_titles}
\end{figure}

Figure~\ref{fig:ground_titles} shows how close the approach can optimize the percentage of ground tiles towards a certain targeted distribution. The results demonstrate that in almost every run we can get very close to a targeted percentage. 

\begin{figure*}[htbp]
\centering
\includegraphics[width=1.0\textwidth]{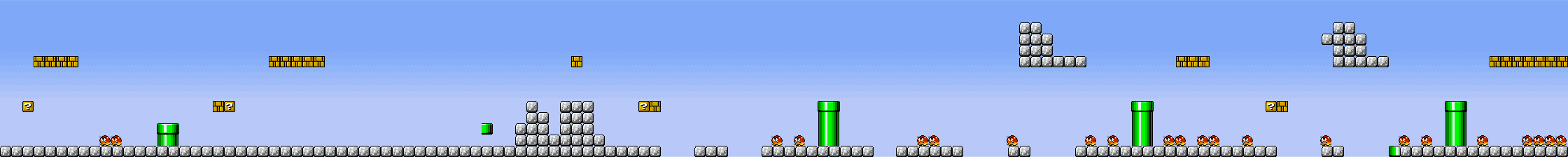}
\caption{Level with increasing difficulty. \normalfont Our LVE approach can create levels composed of multiple parts that gradually increase in difficulty (less ground tiles, more enemies). In the future this approach could be used to create a level in real-time that is tailored to the particular skill of the player (dynamic difficulty adaptation). }
\label{fig:increasing_difficulty}
\end{figure*}

Figure~\ref{fig:increasing_difficulty} shows a level that was created with increasing difficulty in mind: 100\% ground coverage for sections 1 and 2, 70\% coverage for sections 3--5, and maximizing the total number $n$ of enemies for section 4 and 5. The approach is able to optimize both the ground distribution as well as the number of enemies. 

\subsection{Agent-based testing}

\begin{figure*}[t]
\centering
	\begin{subfigure}[b]{0.49\textwidth}
\includegraphics[width=1.0\textwidth]{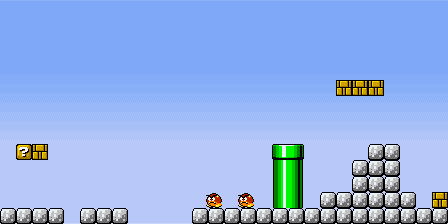}
\caption{Playable level maximizing jumps}
\end{subfigure}
	\begin{subfigure}[b]{0.49\textwidth}
\includegraphics[width=1.0\textwidth]{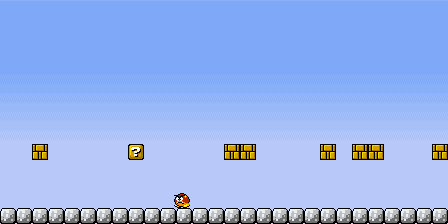}
\caption{Playable level minimizing jumps}
\end{subfigure}
	\begin{subfigure}[b]{0.49\textwidth}
\includegraphics[width=1.0\textwidth]{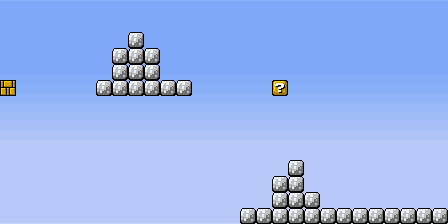}
\caption{Unplayable level}
\end{subfigure}
\begin{subfigure}[b]{0.49\textwidth}
\includegraphics[width=1.0\textwidth]{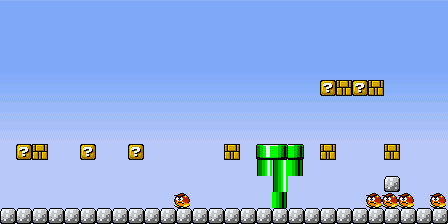}
\caption{Broken titles}
\end{subfigure}
\caption{Agent-based optimization examples. \normalfont (a) and (b) show good examples of levels in which the number of jumps is maximized ($F_{1}$), and minimized ($F_{2}$), respectively. 
%
(c) shows an example of one of the worst individuals found (not playable, $F_{1}$). An example of an individual that reaches high fitness (maximizing jumps, $F_{1}$) but has broken titles is shown in (d). 
%
}
\label{fig:java_examples}
\end{figure*}

Figure~\ref{fig:java_examples} shows some of the best and worst results obtained for both fitness functions. CMA-ES did discover some non-playable levels as depicted in Figure~\ref{fig:java_examples}c. Among the best results for fitness function $F_{1}$ (i.e.\ playable levels with a high number of required jumps) are level sections with and without slight title errors 
(a and d). In the future, the representation of levels could be improved or directly repaired in such a way that the pipes are no longer a cause for visually faulty levels. 
The level depicted in (b) is one of the best examples found when optimizing for  fitness $F_{2}$ (i.e.\ playable with a small number of required jumps). The level requires only one single jump over the enemy and is easy to solve.

\begin{figure}[htbp]
\centering
\includegraphics[width=0.79\linewidth, page=2]{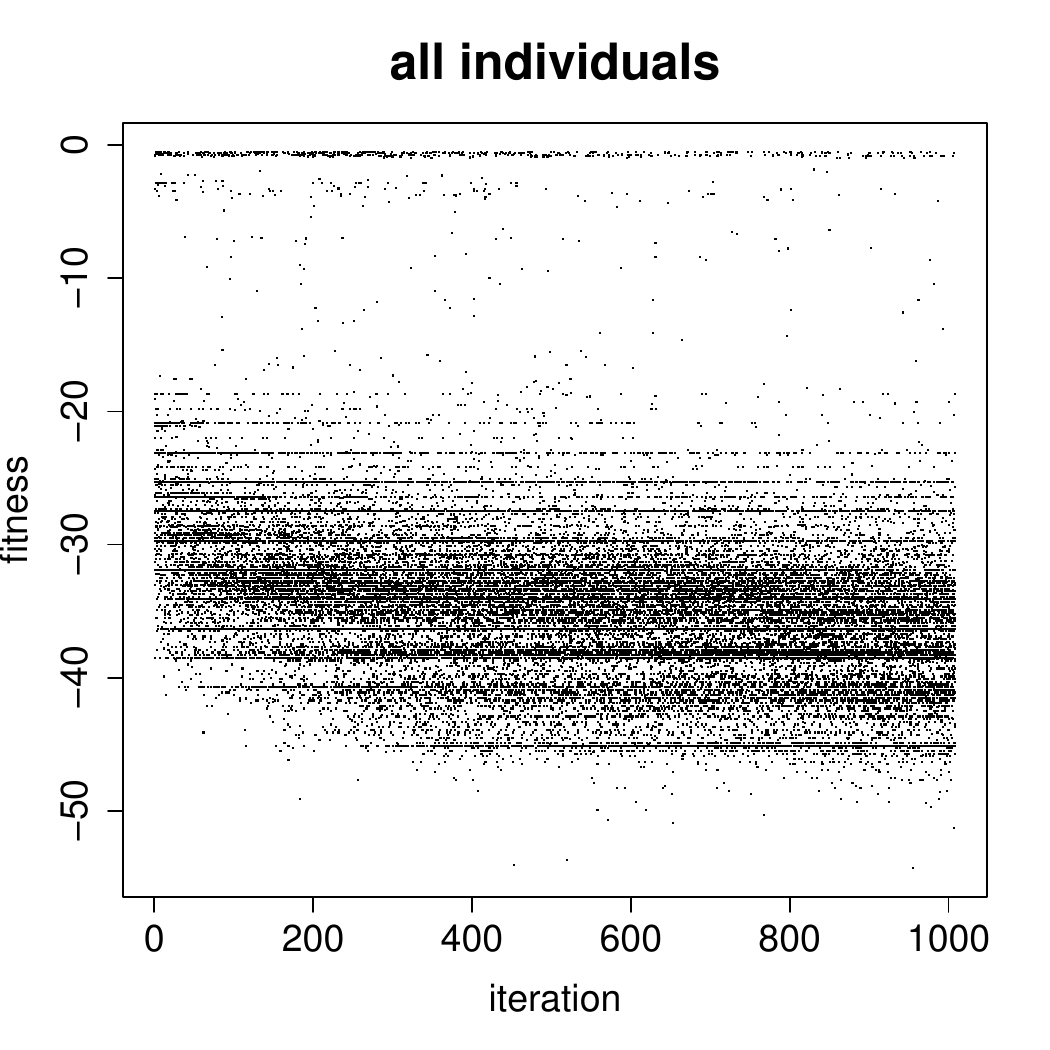}
\hfill
\includegraphics[width=0.79\linewidth, page=3]{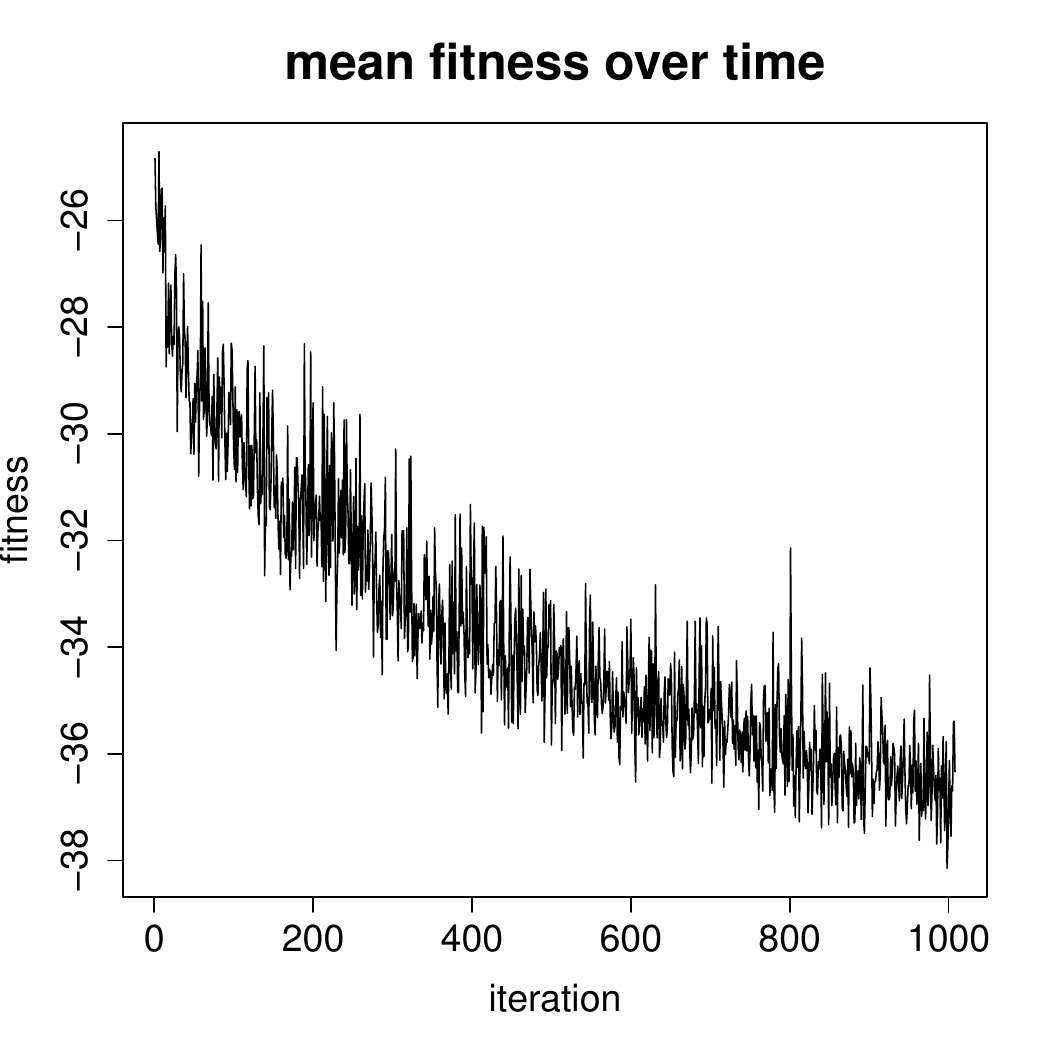}
\caption{\label{fig:timeline}Agent-based fitness progression $F_{1}$: \normalfont Top: Fitness of generated individual at CMA-ES iteration. Bottom: Average fitness of individuals generated at given iteration. Lower values are better.}
\end{figure}

Despite using a noisy fitness function, which is only an approximation of actual level difficulty, the optimization algorithm is able to discover a variety of  interesting results. While we observe some individuals with a small fitness being generated even late into the optimisation process (Figure~\ref{fig:timeline}, top), the average fitness value of generated individuals decreases with increasing iteration (Figure~\ref{fig:timeline}, bottom). The overall decrease of fitness over time does suggest that the GAN-based level generation process is indeed controllable. It is likely that the low-scoring individuals in later iterations result from the fact that levels that require a high jump count and levels that are not playable are close in the search space. We suspect that further modification of the fitness function and using a more robust CMA-ES version intended for noisy optimization could further improve the observed optimization efficiency.

Overall, we show that we are able to create a variety of levels that translate to a plethora of different playthroughs. However, it is of course difficult to find a suitable fitness function, that (1) expresses the desired game qualities but (2) is also tractable for an optimization algorithm. Additionally, the noise of the function should be investigated in depth. Since the evaluation of the fitness function does take considerable time, one should probably also consider using other approaches, for example surrogate-based algorithms.


\subsection{Discussion and Future Work}\label{sec:discuss}
Although GANs are known for their success in generating photo-realistic images (composed of pixels with blendable color values), their application to discrete tiled images is less explored. The results in this paper demonstrate that GANs are in general able to capture the basic structure of a Mario level, i.e. a traversable ground with some obstacles (cf. Figure~\ref{fig:gaussian}). Additionally, we are able to evolve levels that are not just replications of the training examples (compare Figures~\ref{fig:trainingsample} and \ref{fig:increasing_difficulty}).

However, sometimes certain broken structures in the output of the GAN are apparent: e.g.\ incomplete pipes. In the future, this might be addressed by borrowing ideas from text (symbol sequence) generation models such as LSTMs \cite{kusner2016gans}. In these models the discrete choice of symbol at each observable location is conditioned not only on the continuous output of a hidden layer but also the discrete choice of the immediately preceding symbol. This approach would combine the discrete context dependence of Snodgrass' Multi-dimensional Markov Chains (which accurately capture only local tile structures) with the global structure enforced by the upsampling convolutional layers used in our GAN.

An intriguing future possibility is to first train a generator off-line and then 
distribute the architecture and weights of this network with a game so that extremely rapid on-line level generation can be carried out with the model (perhaps to support evolving player-adapted level designs). Depending on the fitness function chosen, this could be employed for both dynamically adapting the difficulty of levels, but also for providing more exploration-focused content by adding more coins in places that are difficult to reach.


Our generator focuses on recreating just the tile-level description of a level primarily because this is the data available in the Video Game Level Corpus. With a richer dataset capturing summaries of player behavior (which actions they typically took when their character occupied a given tile location), we could also train a network to output level designs along with design annotations capturing expectations about player behavior and experience for the newly generated level. Even if these annotation layer outputs go unused for generated levels, having them present in the training data could help the network learn patterns that are specifically relevant to player behavior, beyond basic spatial tile patterns. In general, training with a larger level corpus could allow the GAN to capture a greater variety of different Mario level styles. 

One potential area of future work is the use of Multi-Objective Optimization Algorithms \cite{deb:tec02} to evolve the latent vector using multiple evaluation criteria. Many different criteria can make
video game levels enjoyable to different people, and a typical game
needs to contain a variety of different level types, so evolving
with multiple objective functions could be beneficial. Given such functions, it would also be interesting to compare our results with other procedurally generated content, as well as manually designed levels, in terms of the obtained values. However, further work on automatic game evaluation is required to define purposeful fitness functions. 



\section{Conclusion}\label{sec:conc}
This paper presented a novel latent variable evolution approach that can evolve new Mario levels after being trained in an unsupervised way on an existing Mario level. The approach can optimize levels for different distributions and combinations of tile types, but can also optimize levels with agent-based evaluation functions. While the GAN is often able to capture the high-level structure of the training level, it sometimes produces broken structures. In the future this could be remedied by applying GAN models that are better suited to the discrete representations employed in such video game levels. The main conclusion is that LVE is a promising approach for fast generation of video game levels that could be extended to a variety of other game genres in the future.

\subsection*{Acknowledgements}
The authors would like to thank the Schloss Dagstuhl team and the organisers of the Dagstuhl Seminar 17471 for a creative and productive seminar.

\balance
\bibliographystyle{ACM-Reference-Format}
\bibliography{mariogan}

\end{document}